\documentclass[lettersize,journal]{IEEEtran}
\usepackage{amsmath,amsfonts}
\usepackage{algorithm}
\usepackage{array}
\usepackage[caption=false,font=normalsize,labelfont=sf,textfont=sf]{subfig}
\usepackage{textcomp}
\usepackage{stfloats}
\usepackage{url}
\usepackage{verbatim}
\usepackage{graphicx}
\usepackage{xcolor}
\usepackage{amssymb}
\usepackage{tabularx}
\usepackage{algpseudocode}
\usepackage{dsfont}
\usepackage[sorting=none]{biblatex}
\begin{document}

\title{Breakdown of Conversational Involvement: A Joint Multi-Modal Analysis of Conversation Floors and F-formation Detection with Multiple Features}

\author{Zonghuan Li, Chirag Raman, Hayley Hung~\IEEEmembership{Member,~IEEE} 

\thanks{Zonghuan Li, Chirag Raman and Hayley Hung are with the Intelligent Systems, Delft University of Technology EEMCS, 2628 XE Delft, Zuid-Holland, The Netherlands. Email: \{z.li-25, c.a.raman, h.hung\}@tudelft.nl.}
\thanks{This work is funded by ERC Consolidator Grant NEON:101089163. }%
\thanks{This paper is an extension of the work titled as \textit{Towards automatic estimation of conversation floors within F-formations} presented at
2019 8th International Conference on Affective Computing and Intelligent Interaction Workshops and Demos (ACIIW), Cambridge, UK, September 2019, doi: 10.1109/ACIIW.2019.8925065.}%
\thanks{Corresponding author: Hayley Hung.}
}

\markboth{Preprint.}%
{Li \MakeLowercase{\textit{et al.}}: Breakdown of Conversational Involvement: A Joint Multi-Modal Analysis of Conversation Floors and F-formation Detection with Multiple Features}


\maketitle

\begin{abstract}
Interactions happen universally and spontaneously in daily lives. Studying involvement in interactions helps us analyze who can influence who and has a wide potential in applications. Specifically, we consider conversational involvement in free-standing conversations. We study two operationalizations of an FCG: F-formations and conversation floors, and propose that current assumption that F-formations capture single conversations is insufficient. To illustrate how different implementations could impact the detected F-formations, we employ three analyses. First, we propose that speaking status can be used for estimating conversation floors, and leveraged it to analyze floors within F-formations. Second, we analyze spatial characteristics of the detection results by applying different feature sets on chosen baselines, and find that groups resulting from head features are more likely to be subsumed by groups detected using other features. Finally, we do a joint analysis which combines the spatial features and floor patterns in the conversation. We observe that certain pipelines can detect groups with single floors from upper body features, while retaining F-formations with lower body features. We conclude that a more detailed theory and framework is needed to holistically demonstrate conversational involvement within interactions.
\end{abstract}


\section{Introduction} \label{sec:intro}

As humans, interactions are an indispensable part of our lives. We interact with the environment, with other people, and with artificial systems/agents. Studying interaction and in particular \textit{who can influence who}, is not just a conceptual problem, but also motivated by technical applications. In modern conversational agents, this helps us develop personalized supportive technologies (synchronous or asynchronous, embodied or disembodied), and enhance Human-Robot Interaction (HRI) by identifying the influential source or agents. In fields like health or security, a deeper understanding of who has the influence can help supportive surveillance, for instance, identifying gatherings and detecting potential crimes. Another major application of this is to support people in spontaneous conversations in the wild, which has been linked with organizational learning and entrepreneurial success. 

It is common to define and analyze \textit{involvement} to describe who is influencing who. Multiple sources have linked involvement with attention, coordination, and elaboration~\cite{bohus2009learning, coker1987nature, peters_poggi}. Indeed, involvement can largely represent influence in interactions, and provides crucial insights in interpreting intentions and social experience.

A large part of daily interactions are associated with conversations, and we are in particular interested in those where people can freely move around and forming groups. In such scenes, encapsulating involvement requires knowing who is talking with who, and specifically, needs to know free standing conversation groups, or FCGs. By its notion, it is natural to expect that each FCG contain exactly one conversation. However, conversation itself is complicated, especially in the \textit{temporal} dimension. Indeed, it is hard to delineate the boundary between one conversation and another one, especially with the presence of multiple people joining/leaving, and the change in topics. As we discuss later, this often lead to multiple choices of operationalization of turn, topic, or floor~\cite{edelsky1981s}, and often needs cognitive analysis. 


However, such cognitive analysis is hard to be built upon for automated implementations, and as a result, FCGs are typically operationalized differently. Specifically, \textit{spatial} factors like positions and orientations, are much easier to obtain and compute. Kendon~\cite{kendon} has proposed the term \textit{F-formation} to describe spatial structure of people's interactions, which has then received much attention in the domain of automatic FCG detection. It becomes natural, then, to question whether these FCGs detected as operationalized by F-formations really contain single conversation floors. If the detected F-formations do not align well with the temporal characteristics exhibited by a floor, this implementation is insufficient in terms of accurate locating the conversation. 


In this article, we distinguish but also link the two operationalization mentioned: the conversation floor based on topic analysis and F-formation widely used in FCG detections. We explore the influence of spatial factors on the detection of FCGs and link the results back to what we observe by focusing on floor as the temporal factor. We choose speaking as the decisive indicator of a conversation. Concretely, we pose the following research questions:

\begin{enumerate}
    \item Can we use observed speaker turns to infer the conversation floors within an F-formation?
    \item What is the influence of different spatial factors to detected F-formations in terms of conversation floors?
\end{enumerate}

\noindent Our contributions are as follows: 
\begin{enumerate}
    \item In conceptual perspective, we propose to use speaking turn as an indicator of distinct conversation floors. We also connect this notion to analyzing schisming in conversations~\cite{sacks1974simplest, goodwin1987forgetfulness, egbert1997schisming}.
    \item In methodological perspective, we provide evidence that multiple conversations can exist in detected F-formations, and show that different spatial factors contribute differently in terms of floors. We also provide two metrics to measure spatial relationship within schisming.
\end{enumerate}

The remaining parts of the paper are structured as following. We describe some of the spatial and temporal notions and factors used to study FCGs in section~\ref{sec:background}. In section~\ref{sec:related_work} we provide a literature review involving the use of these spatial or temporal notions, motivating the need to consider both of these aspects. In section~\ref{sec:method}, we first propose to use speaking turns to indicate distinct conversation floors. Then, we propose the Soft Homogeneity (SH) and Split Factor (SF) to analyze spatial relationship between group detection results. Third, we link floor characteristics with different features by performing a joint analysis. In section~\ref{sec:experiments} we provide details of the experimental settings and analyze the results. Finally, section~\ref{sec:conclusion} summarizes our findings and concludes the paper.

\section{Background} \label{sec:background}

\subsection{Spatial factors}
Originally proposed by Kendon~\cite{kendon}, the notion of \textit{F-formation} has become the embodiment of FCG that is adopted in many studies. An F-formation is defined as follows:

\begin{quote}
    An F-formation arises whenever two or more people sustain a spatial and orientational relationship in which the space between them is one to which they have equal, direct, and exclusive access. \cite[p.~150]{kendon}
\end{quote}

Kendon argues that activity is always located, and denotes the space extending in front of a person that is used for the activity as the person’s \textit{transactional segment}. ``When two or more people come together to perform some activity, they are liable to arrange themselves such that their individual transactional segments overlap to create a joint transactional space.'' This joint space between the participants is an \textit{o-space}, and an F-formation arises whenever there is an o-space.

Since this notion depends on physical aspects, there has been many approaches trying to detect F-formations in an automatic way. While at the same time, most of these computational works often assume that the transaction involves a single conversation.


\subsection{Temporal factors}
The conversation of focus in an FCG, however, is dynamic in nature. If conversations change over time, what are the temporal units that describe their underlying structure? Some of the terms used in early literature to organize conversational units include \textit{turn}, \textit{topic}, \textit{gap}, and \textit{floor}. First proposed by ~\cite{Sacks1972c}, Sacks equaled the term of a turn and a floor. On the other hand, Edelsky~\cite{edelsky1981s}, while providing an excellent review of these concepts, has proposed that a floor is different from a turn. She states that most of these units were defined previously on the basis of some technical or mechanical structure such as signals of speakers or listeners, ignoring the intention of the participant. In contrast, Edelsky defines floor from a cognitive perspective:

\begin{quote}
    The floor is defined as the acknowledged what’s going-on within a psychological time/space. What’s going on can be the development of a topic or a function (teasing, soliciting a response, etc.) or an interaction of the two. It can be developed or controlled by one person at a time or by several simultaneously or in quick succession.\cite[p.~405]{edelsky1981s}
\end{quote}

Many later works also adopt to leverage cognitive analysis for defining floors. McNeil~\cite{mcneill2009mind} used the term ``growth point'' to be the mental package that carries linguistic components, and sharing the same growth point indicates that they are within the same mental being. Similarly, ~\cite{hayashi1991floor} proposed that a floor to be ``a means of communicative attention orientation which exists not at the level of turn and move but at a higher level of conversation structure'', which approaches also from a cognitive perspective.


\section{Related Works} \label{sec:related_work}


\subsection{Estimating Involvement}
Involvement, as described by Goffman~\cite{goffman2008behavior}, refers to ``sustain some kind of cognitive and affective engrossment in it, some mobilization of one's psychobiological resources''. Many works have been developed to analyze involvement through verbal cues like turn-taking patterns. Bohus and Horvitz~\cite{bohus2014managing} propose a self-supervised method for forecasting disengagement with an interactive robot using a conservative heuristic. Some works also used turn-taking features to estimate some notion of involvement. Pentland, Madan, and Gips~\cite{pentland2005perception} measured engagement by the z-scored influence each person has on the other’s turn-taking for a pair of participants. Hung and Gatica-Perez~\cite{hung2010estimating} found that the pause duration between an individual’s turns, aggregated at group level, is highly predictive of cohesion in small group meetings. 

However, as the floor is typically held by only a single speaker at a time~\cite{sacks1974simplest} in a conversation, estimating involvement in terms of turns and the floor holder poses potential risk. This is because there is an implicit assumption that involvement is only embodied in speaking behavior. As such, many other studies also tries to detect involvement with silent participants. Oertel, Funes Mora, Gustafson, and Odobez~\cite{oertel2015deciphering} characterize silent participants into multiple categories (attentive listener, side participant, bystander) from audiovisual cues. Oertel and Salvi~\cite{oertel2013gaze} also show that it is possible to estimate individual engagement and group involvement in a multiparty corpus by analyzing the participants’ eye-gaze patterns. 

The following works demonstrate that the task of estimating participant involvement is subjective in nature. Zhang and Hung~\cite{zhang2016beyond, zhang2018social} study the task of detecting associates of an F-formation: members that are attached to an F-formation but do not have full status. They argue that the labeling of conversation groups is not an objective task. By collecting multiple annotations of perceived associates, they demonstrate how detecting them can improve initial estimates of full-members of an F-formation. 



\subsection{Detecting FCGs}
In most works regarding the detection and locating the conversations, an FCG is operationalized as an F-formation. Early work on the task of detecting FCGs in video data developed concurrently from two perspectives: those that estimate the location of the o-space~\cite{cristani2011social, setti2013multi}; or those that view an F-formation as a set with individuals being assigned exclusive membership~\cite{dominant_sets, setti2015f}. There has also been considerable work focused on incorporating temporal information for the same task of detecting conversational groups~\cite{tran2014activity, alameda2015analyzing, ricci2015uncovering, GTCG}. Notably, these approaches utilize the head pose as a proxy for the transactional segment (in addition to the body pose) to model F-formation membership. The head pose also acts as a proxy for the Visual Focus of Attention (VFoA)~\cite{bazzani2013social} which further perpetuates an implicit assumption that the joint mutual gaze exists and must be singular within an F-formation. Vazquez, Steinfeld, and Hudson~\cite{vazquez2015parallel} also assume a single conversation within an F-formation while developing a policy for a robot to be aware of a single focus of attention of the conversation. Recently, methods leveraging Visual Language Models (VLMs) were proposed to predict conversation groups dynamically~\cite{yokoyama_dynamic_2025}. However, it does not explicitly distinct in the notion between F-formations and conversation floors.

\subsection{Schisming in conversation}
For a single conversation, it has been pointed out that it is hard to keep more than 4 participants. In a conversation with at least four participants, sometimes it splits up into two or more conversations. This transformation is referred to as a schism~\cite{egbert1997schisming} or schisming. 

One of the earliest allusions to the phenomenon of schisming based on anecdotal evidence occurs in the work of Goffman, who suggested that a gathering of two participants exhausts an encounter and forms a fully-focused gathering~\cite{goffman2008behavior}. With more than two participants, there may be persons officially present in the situation who are not themselves so engaged. These bystanders change the gathering into a partly-focused one. If more than three persons are present, there may be more than one encounter carried on in the same situation, resulting in a multi-focused gathering, which effectively has introduced schisming. In subsequent work, Sacks, Schegloff, and Jefferson~\cite{sacks1974simplest} and Goodwin~\cite{goodwin1987forgetfulness} both indicated that the co-existence of two turn-taking systems is the most decisive characteristic of schisming. Specifically, Egbert also makes a systematic differentiation between overlap and simultaneous talk during schisming. In overlap, simultaneous speakers compete for the floor, an event usually resolved by returning to one-speaker-at-a-time. In schisming by contrast, simultaneous speakers orient to one of two distinct floors, an event which if resolved successfully, results in the establishment of two floors~\cite{egbert1997schisming}. Later, Dunbar, Duncan, and Nettle~\cite{dunbar1995size} demonstrates ``clique size'' which describes the number of people in a conversation. It reveals that the frequency of occurrence drops significantly for cliques with size equal or above 4. 

Other studies attempts to answer the ``why'' question to this limitation. More recently, ~\cite{cowan2010magical} has attributed it to limited cognitive capacity, which could store only 3 to 5 meaningful items for adults. The author further hypothesizes that too many working memory items could prohibit the brain from keeping all of them active, assuming a turn-taking pattern in neuron firing. ~\cite{krems2019conversations} attempts to answer from a societal perspective, by experimenting with mathematical models that simulate human society and count the pairs of inclusive dyads (including the particular person) and exclusive dyads (excluding the particular person). Since the number of inclusive dyads grow linearly and the number of exclusive dyads grows quadratically, they propose that it is possible that people benefit more from being in the inclusive dyads and not in the exclusive ones, and thus inhibits forming conversations of more than 4 people.





\section{Method} \label{sec:method}

In this section we build upon the previously discussed concepts to distinguish as well as link the spatial and temporal factors in F-formation detection. First, we propose to use simultaneous speakers in an F-formation as an indicator of the existence of distinct conversation floors. Second, we relate the detected F-formations with different body clues in spatial aspect with SH and SF metrics. Finally, we propose a joint analysis method to combine the spatial relationship with temporal floor pattern on the detections.

\begin{figure}
    \centering
    \includegraphics[width=\linewidth]{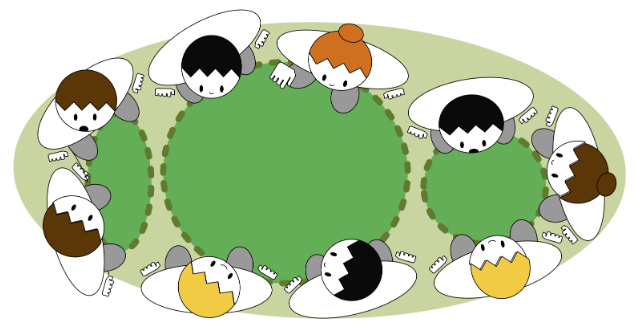}
    \caption{Depiction of a single F-formation with multiple conversation floors. The darker green regions within dotted lines represent distinct simultaneous conversation floors. Most works representing a conversing group as an F-formation make the simplifying assumption that a single conversation occurs within an F-formation with a joint focus of attention for all members.}
    \label{fig:floor-fformation}
\end{figure}

\subsection{Analyzing temporal characteristic of an F-formation}

Conversations are inextricably connected with the content of speech. However, a common concern with observing conversation groups by verbal information is the potential violation of privacy. How can we then investigate the existence of distinct conversations without high-quality audio (verbal data)? We propose the following insights:

\subsubsection{Inferring schisms without audio data}
The relationship between body movements such as gestures and aspects in speech such as emphasis, rhythm, and content has been long established in literature~\cite{bull1985body, dittmann1969body, mcneill2000language}. Some works have shown promising results in estimating the presence of voice activity from automated gestural analysis or accelerometer data~\cite{hung2010speech, gedik2016speaking, gedik2017personalised, gedik2018detecting}. These approaches that refrain from using privacy-sensitive data suggest that it is feasible to automatically estimate speaking behavior, and in turn, speaking turns, without audio. Additionally, since the co-existence of two turn-taking systems is the most decisive characteristic of schisming, we argue that it is in turn reasonable to explore the inference of schisms without audio data, but through speaking turns.

\subsubsection{Linking schisming to floors and F-formations}

Schisming, by its nature, influences floors as Edelsky defined as the ``acknowledged what’s-going-on within a psychological time space''. This acknowledgment is essentially a focus, either on topic or some other functions. When schisming occurs, it changes this single-focused interaction into a multi-focused one, and each of the focused gathering is a new floor.

When thinking of the influence of schism to F-formations defined by Kendon, on the other hand, we find that it may not essentially change it. Kendon defines F-formations based on transactional segments, which come from lower-body orientations. If the participants moved their upper-bodies to form multiple floors, while the lower bodies remain configured such that an common o-space is still maintained, then they form multiple conversation floors within the same F-formation. This phenomenon is shown conceptually in figure~\ref{fig:floor-fformation}. 

\begin{figure}
    \centering
    \includegraphics[width=\linewidth]{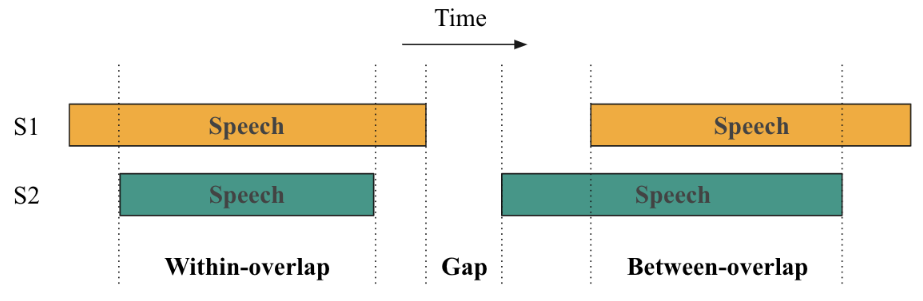}
    \caption{Illustration of gaps, within-overlaps, and between-overlaps for two speakers (S1 and S2) within the same floor. The scheme was originally proposed by Heldner and Edlund~\cite{heldner2010pauses} and adopted by Levinson and Torreira in their analysis~\cite{levinson2015timing}.}
    \label{fig:speaker-duration}
\end{figure}

Combining the discussions, we argue that it is feasible to detect distinct floors within an F-formation without audio. Specifically, we consider the following metric: for a time window $w$ with duration $d$, we define a person to be a speaker if they speak for the entire period. For a group with multiple persons, the number of floors within it for $w$ is then the number of \textit{simultaneous} speakers, since we are inferring floors from speaking turns, and each speaker corresponds to a turn holder.

\subsubsection{Choice of speaking window duration}

In order to capture useful dynamics in a conversation about floors, we need to choose proper $d$s. As in Egbert's study, short overlapping speech can exist within one floor. Thus, if the duration was too short, it may capture such back-channels of overlaps, and a floor thus detected is not a sensible one. If the duration was too long, the simultaneous speaker number may quickly drop to 0, as having just 2 people speaking for an extremely long period of time in a group is very unlikely. Therefore, setting an upper and lower bound for $d$s and vary them in the range might be a good choice. In this article, we vary $d$ from 1 second to 20 seconds with an interval of 1 second.

Our choice of the duration comes from related studies. Heldner and Edlund studied gaps and overlaps in conversations, and discovered that on average 40\% of the speaker transitions contain overlaps ~\cite{heldner2010pauses}. In Spoken Dutch Corpus they used, their histogram shows that the duration of these overlaps follows has a mean of 610 ms and median of 470 ms, all under one second. Levinson and Torreira differentiate between types of overlaps in a later research: 1) between overlaps, which refer to overlaps where the floor holder was changed, but without a silent gap between speakers; 2) within-overlaps, where the floor holder was unchanged, and overlapping speech occurred within a speaking turn~\cite{levinson2015timing}. These two types are illustrated in figure~\ref{fig:speaker-duration}. For the duration, between overlaps had a modal duration of 96 ms, a median of 205 ms, a mean of 275 ms. On the other hand, within-overlaps exhibited an estimated modal duration of 350 ms, a median of 389 ms, a mean of 447 ms. Within the same study, they also found that only 3.8\% of the time were simultaneous speaking from two speakers observable, by using Switchboard Corpus of English telephone conversations. This aligns with Sacks et al’s conclusion that “overwhelmingly, one party speaks at a time”~\cite{sacks1974simplest}. 

These results suggest that although back-channels and overlapping speech is common, the most of them are rather short and do not indicate a change in floors. Therefore, we can choose the lower bound of window duration $d$ such that it filters out those overlaps that do not come with a change in floors. For the upper bound, a reasonable value should be at least greater than the average turn duration of a speaker. As in~\cite{levinson2015timing}, Levinson and Torreira report that contiguous speech delimited by a silent interval of at least 180 ms had a mean duration of 1680 ms and a median of 1227 ms. Following this, we leverage a 1 second minimal duration for analysis.

\subsection{Combining turn-taking and spatial relationship within F-formation detections}


\subsubsection{Using different sets of features for F-formation detection}


Although Kendon has described F-formations based on lower-body orientations, he also mentions that one may move their head or upper body parts to adapt to conversations without moving their lower bodies: 

\begin{quote}
    However, if \textit{p} rotates his head so that a line projected from the midline of his face forms an angle of more than thirty degrees from the midline of his lower body, \textit{p} may be said to be facing out of his transactional segment. It is not uncommon for someone to look out of his transactional segment in this way – for example when someone looks over his shoulder. Such head orientations are rarely continued for long, however. Any sustained looking in such a direction is usually associated with a reorientation of the lower body, so that the direction of the transactional segment again coincides with the direction in which the face is oriented.~\cite[p.~212]{kendon}
\end{quote}

Therefore, moving the head away from the transactional segment is not uncommon, although they typically don't last for very long. Also, this aligns exactly with our previous discussion that one may move their upper body towards conversations while still remaining configured in an F-formations with the lower bodies. This naturally raises the following question: if there exists F-formation detections that leverage the head or body orientation~\cite{GTCG,GCFF}, can we detect this upper/lower body spatial distinction with multiple clues from the body? Specifically, with the same detection pipeline, can we observe the pattern that groups resulting from lower body features (LBF) correspond more to F-formations, while those from upper body features (UBF) correspond more to single conversation groups? This enables us to extract the features (position and orientation data) from different body parts and leverage them in existing pipelines. In this paper, we utilize features from 4 body parts: $\mathcal C=\{\text{head, shoulder, hip, foot}\}$.

\begin{figure*}[!t]
    \centering
    \includegraphics[width=\linewidth]{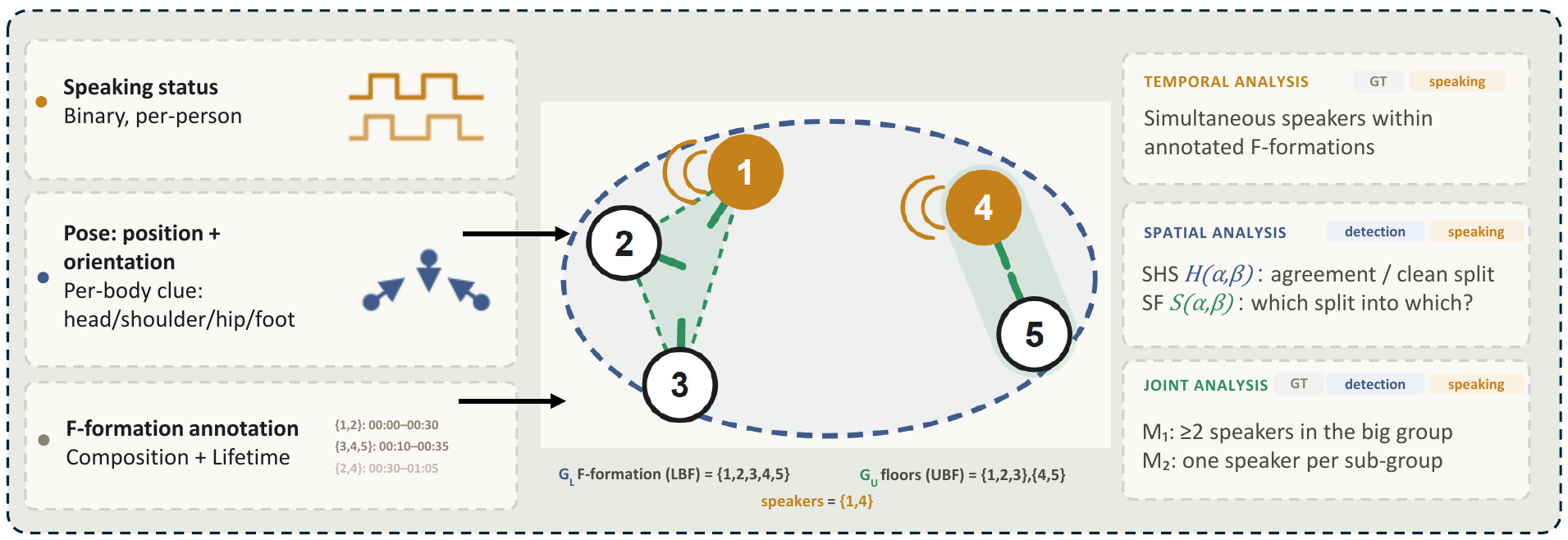}
    \caption{Illustration of the experimental design. We start with a dataset with speaking status and F-formation annotations, as well as body spatial information of human bodies. Temporal analysis: we analyze simultaneous speakers within an F-formation to demonstrate there might be more than one speaker; Spatial analysis: We run selected F-formation detection pipelines and analyze spatial patterns between them; Joint analysis: we combine the spatial features with conversation floors to study the relationship between detection results and actual physically embodied floors.}
    \label{fig:pipeline}
\end{figure*}
\subsubsection{Measuring spatial relationships}

We argued earlier that simultaneous floors could exist within an F-formation. From the spatial perspective, we operationalize this as a \textbf{subsumption relationship} between groups. To be more specific, given a whole set of people $P$, which represents every participant in a scene or frame, a \textit{group} $g$ is a set of people. Running an F-formation detection baseline on one image with people $P$ would give a \textit{grouping} $G=\{g_1, \dots, g_k\}$, which is a partition of the entire set $P$ into one or more smaller groups, where 
\begin{equation}
g_i \cap g_j = \varnothing, \quad \forall i \neq j, \quad \text{and} \quad \bigcup_{i=1}^k g_i = P
\end{equation}
i.e. any two different subsets are disjoint and the union of all subsets is the whole set $P$. As previously stated, we are interested in the case where people in the same conversation floor rotate their heads towards each other, while their lower bodies remain configured in the same F-formation. This is embodied in the spatial relationship that groups detected from LBF could split into groups detected from UBF. In other words, given the grouping $G_U$ from UBF and $G_L$ from LBF, 
\begin{equation} \label{eq:group_subsumption}
    \exists g_L \in G_L, \exists g_U^{(1)},g_U^{(2)} \in G_U, \quad  \text{s.t.} \quad g_U^{(1)},g_U^{(2)} \subsetneq g_L
\end{equation}
i.e. there exists a group in $G_L$ such that at least two groups in $G_U$ are its proper subset.

However, for a dataset with F-formation annotations, there are several issues with the direct measurement of different detected groups:
\begin{itemize}
    \item The detections are usually unstable. For many methods that utilize body positions and orientations to infer F-formations, neighboring frames can have different results, although they vary little in terms of the positions and orientations. Our analysis, on the other hand, needs temporal consistency to analyze the conversation floors. The annotations, on the other hand, provides a good anchor for this as they are stable and have a clear, tractable lifetime.
    \item The annotations may or may not correspond to true F-formations. Many datasets provide ``F-formation'' or ``conversation group'' annotations, but since they typically do not distinct one from the other, it is unknown whether the annotations are made by leveraging upper or lower body orientations. In fact, we have observed many instances where the annotation actually aligns more with the conversation groups by looking at upper body orientations. Therefore, anchoring the annotations as the ground truths of F-formations may lose the larger context.
\end{itemize}

With the above issues, we design the following analysis pipeline. We start from a dataset with F-formation annotations $\{A_i\}_{i=1}^{N_A}$. Each annotated F-formation is a group with its composition $A_i$ and lifetime $\mathcal{T}_i= \{t_{1}^{(i)}, \dots, t_{T_i}^{(i)}\}$. For each frame $t$, we have the groupings detected using different feature sets $\{G_\alpha^{t}\}_{t=t_1^{(i)}}^{t_{T_i}^{(i)}}$ where $\alpha \in \mathcal{C}$. Within each frame, we define the \textbf{relative detection groups} with respect to $A_i$ as:
\begin{equation}
    \mathcal{R}_\alpha(A_i) = \{g| g\in G_\alpha, g\cap A_i \neq \varnothing \}.
\end{equation}
Since $A_i$ may align more with F-formations or conversation floors, this allows us to have a full contextual picture of people potentially in the same F-formation, regardless of whether $A_i$ is closer to either distinction.

\begin{figure}
    \centering
    \includegraphics[width=\linewidth]{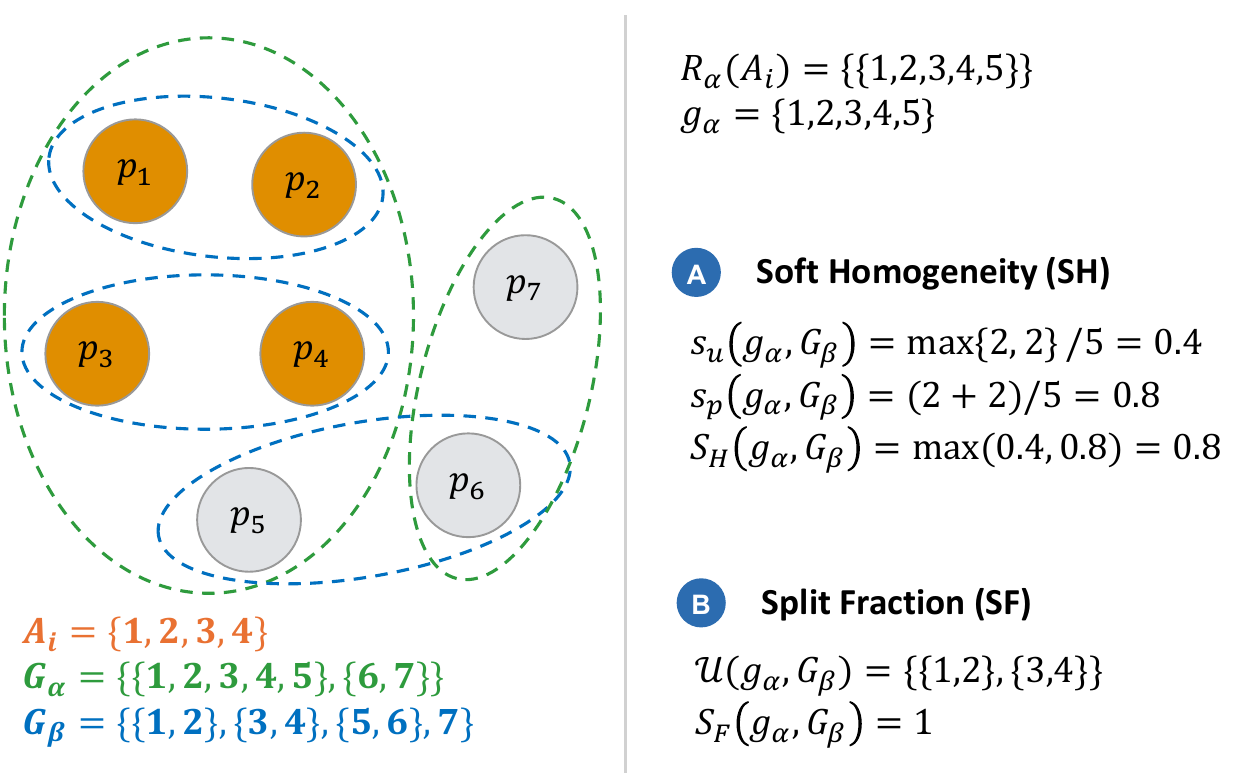}
    \caption{Illustration of spatial measures $S_H$ and $S_F$. First, we identify the composition of annotation $A_i$, detection results $G_\alpha,G_\beta$, and the relative detection $\mathcal{R}_\alpha (A_i)$. For SH, we first calculate $s_u$ and $s_p$, then take the maximum to get $S_H$. For SF, it is $1$ when there are at least 2 subsets of $g_\alpha$ in $G_\beta$ with at least 2 members, and $0$ otherwise.}
    \label{fig:spatial_illustration}
\end{figure}

With each pair of $\alpha, \beta \in \mathcal{C}$ and each frame detection $G_\alpha$, we compute the following asymmetric pairwise metrics:

\noindent \textbf{Soft Homogeneity (SH).} Inspired from the entropy-based homogeneity score~\cite{rosenberg2007v} which evaluates the level of homogeneity over clustering, our proposed SH measures whether one group $g_\alpha$ subsumes other groups in $G_\beta$, or the other way around (being subsumed in one group in $G_\beta$). We start with the soft subsumption score $s_u$ and soft partition score $s_p$, and take SH as the maximum among them:
\begin{equation}
    s_u(g_\alpha, G_\beta) = \max\limits_{g_\beta \in G_\beta} |g_\alpha \cap g_\beta| / |g_\alpha|
\end{equation}
\begin{equation}
    s_p(g_\alpha, G_\beta) =\left|\bigcup{g_\beta \in G_\beta : g_\beta \subseteq g_\alpha}\right| / |g_\alpha|
\end{equation}
\begin{equation}
    S_H(g_\alpha, G_\beta) = \max[s_u(g_\alpha, G_\beta), s_p(g_\alpha, G_\beta)]
\end{equation}
For one specific group $g_\alpha$, $s_u$ shows the degree that it is subsumed by other groups in $G_\beta$: it is $1$ if and only if it is completely subsumed by another group in $G_\beta$, and decreases if its members are distributed across different groups in $G_\beta$. $s_p$ shows the degree that smaller groups in $G_\beta$ covers it, and only take to 1 when there it is fully partitioned into smaller groups in $G_\beta$. Finally, $S_H$ takes the larger number between $s_u$ and $s_p$. Intuitively, $S_H(g_\alpha, G_\beta)$ is high when $g_\alpha$ has an identical group in $G_\beta$, or splits into multiple groups in $G_\beta$, or is subsumed by a larger group in $G_\beta$, and becomes low when members in $g_\alpha$ spreads across multiple groups in $G_\beta$. When two detection results agree with each other, or a clean split pattern exists, the SH would be high. On the other hand, if the detections from $\alpha$ and $\beta$ are very irrelevant and produce random results, SH would be low.


\noindent \textbf{Split Fraction (SF).} Although SH accounts for whether the detections from clue $\alpha$ and $\beta$ ``agree'' in the sense of group level splitting, it does not tell which way around. For example, SH could be high because that groups in $G_\alpha$ splits into other groups in $G_\beta$, or inversely, or because that groups in $G_\alpha$ are nearly identical to those in $G_\beta$. To address the feature in equation~\ref{eq:group_subsumption}, we employ SF as follows.

Given annotation group $A_i$ and frame $t$, we define the subsumption subset as
\begin{equation}
    \mathcal{U}(g_\alpha, G_\beta) := \{g_\beta | g_\beta \in G_\beta, |g_\beta| \geq 2, g_\beta \subsetneq g_\alpha\},
\end{equation}
which measures when the outer group $g_\alpha$ from body clue $\alpha$ contains at least two non-trivial sub-groups in the inner clue $\beta$'s detections. Based on it we compute the SF as
\begin{equation}
S_F(g_\alpha, G_\beta) = \mathds{1} \{ | \mathcal{U}(g_\alpha, G_\beta) | \geq 2\}
\end{equation}
which counts whether there are at least two proper subsets of $g_\alpha$ existing in $G_\beta$. This directly address if a splitting pattern of $g_\alpha$ exists within $G_\beta$, and measures the direction of splitting that is ambiguous in SH. It also ignores identical grouping results, letting us focus on the non-trivial splitting patterns. Figure~\ref{fig:spatial_illustration} demonstrates the $S_H$ and $S_F$ scores.

\noindent\textbf{Aggregation.} We compute the aggregated score for both SH and SF by averaging over relative detection groups and then over all frames and all annotation F-formations:
\begin{equation} \label{eq:aggregation_1}
    \Theta(\alpha, \beta; A_i, t) = \frac{1}{|\mathcal{R}_\alpha (A_i)|} \sum_{g_\alpha \in \mathcal{R}_\alpha (A_i)} \Theta(g_\alpha, G_\beta^t),
\end{equation}
\begin{equation} \label{eq:aggregation_2}
    \Theta(\alpha, \beta) = \sum_{i=1}^{N_A} \sum_{t\in \mathcal{T}_i} \Theta(\alpha, \beta; A_i, t) \Big/ \sum_{i=1}^{N_A} |\mathcal{T}_i|.
\end{equation}
where $\Theta \in \{S_H, S_F\}$ represents either score.

\subsubsection{Joint analysis with spatial-temporal characteristics}

\begin{figure*}
    \centering
    \includegraphics[width=\linewidth]{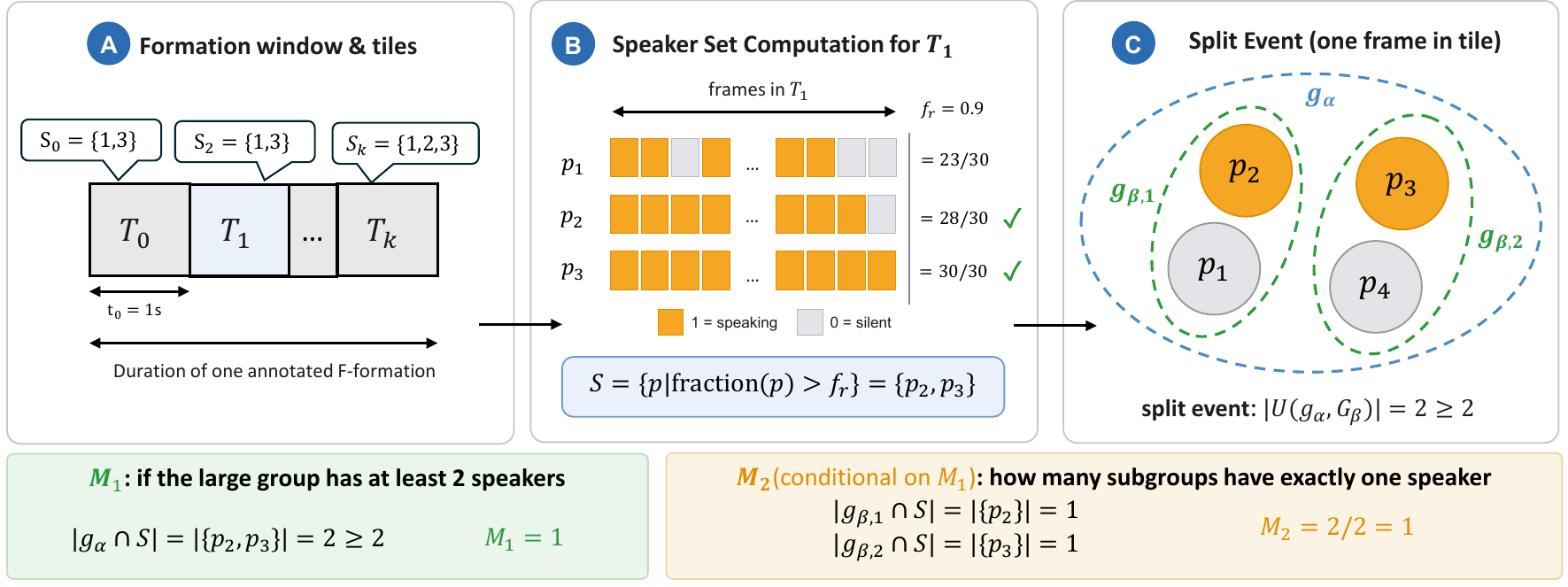}
    \caption{Illustration of joint analysis pipeline. First we split the duration of an annotated F-formation into tiles of length $t_0$, then compute speaker within each tile. For each frame in each tile where a split event exists, we compute $M_1$ and $M_2$. $M_1$ calculates whether the larger group contain at least 2 speakers, and $M_2$ calculates the proportion of subgroups that contain exactly one speaker. Together, they measure how much the detected split events align with the characteristic that multiple floors exist within one F-formation.}
    \label{fig:joint_analysis_pipeline}
\end{figure*}
SH and SF measures whether a splitting pattern exists in the detection results from two body clues $\alpha, \beta$. However, it does directly link back to our observation that more than one floor exist within one F-formation. Concretely speaking, we want to know that if this splitting really corresponds to the F-formation-conversation group distinction illustrated in figure~\ref{fig:floor-fformation}. To demonstrate this, we employ the joint analysis with spatial-temporal characteristics.

We continue to use contiguous speaking status within a window as an indication of floor occupation. Within each annotated F-formation window, we partition time into non-overlapping tiles of duration $t_0$. For each tile, we determine the speaker set $\mathcal{S}$ -- the set of people whose speaking status equals 1 for at least $f_r$ of their annotated frames in the tile. Every frame $t$ inherits the $\mathcal{S}$ of its enclosing tile. We set $t_0=1$s, as it is the shortest window we use for analyzing conversation floors and affords to analyze more F-formations which are short. We choose $f_r=0.9$ to rule out noise in the speaking status annotation.

For each annotated F-formation $A_i$, we define frames $t$ that contain a \textbf{split event} as follows:
\begin{equation}
    \Psi_1^{(i)}(\alpha, \beta) = \{t\in \mathcal{T}_i |\exists g_\alpha \in \mathcal{R}_\alpha (A_i), S_F(g_\alpha, G_\beta^t)=1\}
\end{equation}
and for each $t\in \Psi_1^{(i)}(\alpha, \beta)$ we compute
\begin{equation}
    M_1(g_\alpha, G_\beta) = \mathds{1} \{|g_\alpha \cap \mathcal{S} | \geq 2 \},
\end{equation}
where each $\mathcal{S}$ is computed based on the tile that the frame $t$ belongs to. \textbf{$M_1$ indicates when a split event happens, whether the larger group has at least 2 speakers.}

Built upon $M_1$, we further define all frames with $M_1=1$:
\begin{equation}
    \Psi_2^{(i)}(\alpha, \beta) = \{t\in \Psi_1^{(i)}(\alpha, \beta) |\exists g_\alpha \in \mathcal{R}_\alpha (A_i), M_1(g_\alpha, G_\beta) =1\}
\end{equation}
and for each $t\in \Psi_m^{(i)}(\alpha, \beta)$ we compute metric $M_2$:
\begin{equation}
    M_2(g_\alpha, G_\beta) = \frac{|\{ g_\beta | g_\beta \in \mathcal{U}(g_\alpha, G_\beta), \: |g_\beta \cap \mathcal{S}| =1 \}|}{|\mathcal{U}(g_\alpha, G_\beta)|} 
\end{equation}
\textbf{which measures the fraction of detected sub-groups that are occupied by exactly one speaker -- the ideal signature of one dedicated floor per spatial sub-group.}

\noindent\textbf{Aggregation}. Since the tiles are of equal lengths and non-overlapping, aggregating within tiles first then across tiles is equivalent with aggregating directly over frames. Therefore, we first leverage equation~\ref{eq:aggregation_1} to compute per frame metric for $\Theta \in \{M_1,M_2\}$ and use
\begin{equation} \label{eq:aggregation_3}
    M_j(\alpha, \beta) = \sum_{i=1}^{N_A} \sum_{t\in \Psi_j^{(i)}(\alpha,\beta)} \Theta(\alpha, \beta; A_i, t) \Big/ \sum_{i=1}^{N_A} \left| \Psi_j^{(i)}(\alpha,\beta) \right|.
\end{equation}
to aggregate over all formation, weighted by valid frame numbers where $j\in \{1,2\}$.

\section{Experiments} \label{sec:experiments}

\subsection{Dataset and baseline} \label{sec:dataset_baseline}
Our desiderata of the dataset are in two folds. First, it needs to have the body clues from different parts to allow for clue-wise F-formation detection. Second, having annotations for F-formations and speaking status is essential as they are needed for anchoring spatial analysis and inferring conversation floors respectively. We therefore choose the ConfLab dataset~\cite{raman2022conflab} which contains videos of free-mingling scenes. It has rich scences recorded at high resolution which allows us to extract positions $\mathbf{p}_i$ and orientations $\theta_i$ for each person in each frame. Detailed process of the extraction is provided in the supplementary. Also, it has F-formation and speaking status annotation to facilitate our designed analysis of inferring floors without explicit audio data.


We choose the GTCG~\cite{GTCG} and GCFF~\cite{GCFF} baselines. The former models the transactional segment by sampling points and calculates pairwise affinity matrix by measuring the distances between distributions. The latter  votes for centers of the o-spaces and optimize by assigning and updating people to these centers. Both pipelines provide an explicit and explainable description of the spatial structure of an F-formation by either modeling transactional segments or formation centers, and they provide grouping results $G_\alpha$ with respect to features constructed from feature $\alpha$ for each frame by only taking positions and orientations for each person. 

For both pipelines, the parameters are chosen empirically, and later verified that they are close to optimal choice where the detection results ($F_1$) reaches maximum by averaging all 4 feature set detection results. Specifically, for GTCG we use frustum$=110$cm aperture$=160$ degrees, and for GCFF we use stride$=40$cm and MDL$=6000$.

\begin{figure*}
    \centering
    \includegraphics[width=\linewidth]{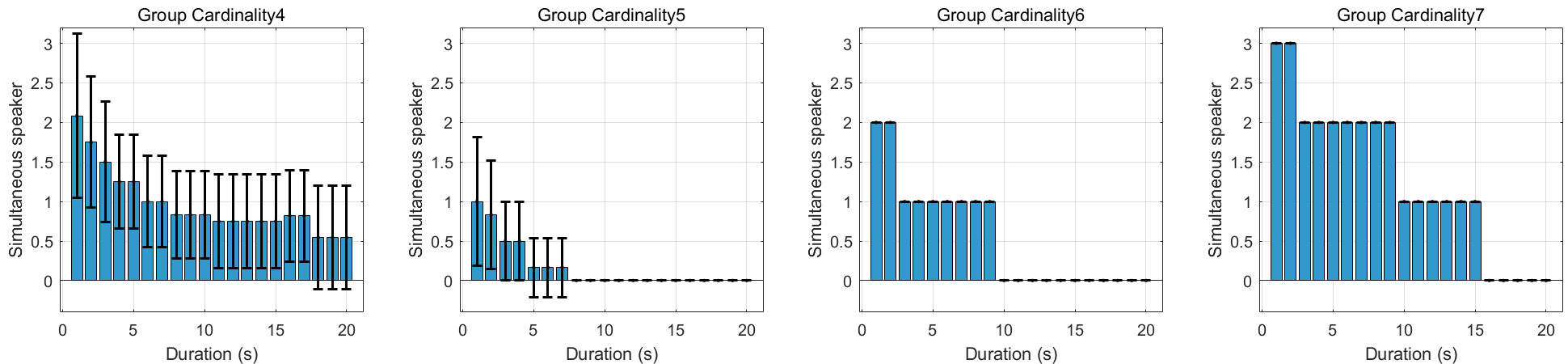}
    \caption{Simultaneous speakers per cardinality of detection using annotations. Each bar is average speaker number with standard deviation, with respect to a specific window duration. For groups with cardinality 4, more than 1 speaker is observed for 1-5 seconds of window. For groups with cardinality 5, no average speaker number is above 1. For cardinality 6-7, although observed with more than 1 speaker for windows duration up to 2 and 9 seconds respectively, the number of samples also drops substantially, resulting in no standard deviation.}
    \label{fig:floor_analysis_gt}
\end{figure*}

\subsection{Turn-Taking Analysis Results}

\subsubsection{\textbf{There is indication that multiple simultaneous speakers exist within annotations of F-formations for ConfLab}} We apply the sliding window varying from 1 to 20 seconds on annotations of ConfLab. We take each continuous sequence of identical annotation as the lifetime of an F-formation. The simultaneous speaker over window length plots are shown in figure~\ref{fig:floor_analysis_gt}. We can observe that simultaneous speaker number decreases as window duration increases, which matches with our expectation that it becomes rarer for cases where people simultaneously speak for longer. For groups of size 4, the simultaneous speaker number reaches around 2 when window size is 1 second, and keeps decreasing but stays above 1 until window duration reaches 6 seconds. For groups of size 5, however, the simultaneous speaker number is generally below 1. For groups of size 6 and 7, we also observe more than one simultaneous speakers in window lengths of below 3 and 10 seconds respectively. The 0 variance in the last 2 plots indicate that there are only one instance for both cases.

We can infer that there is evidence that more than one simultaneous speakers exist within the annotated F-formations in ConfLab. Although for group size of 5 this number is below 1, it could be influenced by the scarcity of samples, and we propose that more cases in group sizes 4, 6, and 7 are persuasive enough to draw our conclusion.

\subsubsection{\textbf{Group cardinality has a significant impact on the turn durations of simultaneous speakers}} 

Previously, we discussed that it is difficult for a single conversation to sustain over four participants. In this experiment, we explore this effect and study the following question: for a given speaking turn duration $d$, do we observe a decrease in the maximum number of conversation floors observed over an F-formation’s lifetime with an increase in the cardinality of an F-formation? Qualitatively, this corresponds to the steepness of fall-off of the curves in~\ref{fig:floor_analysis_gt}. To account for the influence of group cardinality and window duration, we model the simultaneous speaker $y_i$ as instances of independent Poisson variables: $Y_i \sim \log(\mu_i)$, and $\mu_i$ is modeled by the following formula:
\begin{equation}
    \log(\mu_i) = \beta_0 + \beta_1 d_i + \beta_2 c_i + \beta_3 d_i c_i
\end{equation}
where $d_i$ denotes the duration of the window and $c_i$ denotes the group cardinality. The $\beta_i$s are regression coefficients. We treat each possible observation of a window as a data point. For example, an F-formation of cardinality 4 with the lifetime of 20 seconds would result in 20 windows of duration of 1 second. This would add 20 data points with $c_i=4$ and $d_i=1$. $y_i$s for these data would correspond to the number of simultaneous speakers of the respective windows.

We fit the data using a Generalized Linear Model (GLM) by the Python \textit{statsmodel}~\footnote{https://www.statsmodels.org/stable/index.html} package. The results are shown in table~\ref{tab:glm}. We observe that the turn duration and turn-duration:cardinality factors have an impact on the simultaneous speakers at significance level of 0.01, while cardinality alone have an impact with significance level of 0.05.

Although this test tells us that all three factors influence the simultaneous speaker numbers, we still need a post-hoc analysis to confirm each possible pair of cardinality makes difference among them. Therefore, we apply multiple GLMs using the same formula and present the results in table~\ref{tab:glm_pair}. In the last column, we observe specifically that cardinality and its interaction between turn duration are significant at level 0.001 for $\{4/5\}$ pair.

\begin{table}[!t]
    \caption{Generalized model regression results}
    \label{tab:glm}
    \centering
    \begin{tabular}{ccccc}
    \hline
        & \textbf{Coeff}($\beta$) & \textbf{Std Err} & $z$ & $P>z$ \\
       \hline
       \textbf{Intercept} & -0.7164 & 0.564 & -1.270 & 0.204 \\
       \textbf{Turn-duration} & 0.2424 & 0.076 & 3.173 & 0.002 \\
       \textbf{Cardinality} & 0.2726 & 0.121 & 2.244 & 0.025 \\
       \textbf{Turn-duration:Cardinality} & -0.0754 & 0.018 & -4.298 & 0.000 \\
       \hline
    \end{tabular}
\end{table}

\begin{table}[!t]
    \caption{Nominal P-values for six post-hoc GLM regression comparisons}
    \label{tab:glm_pair}
    \centering
    \begin{tabular}{|c|cccc|}
    \hline
        \textbf{Cardinality pairs} & \textbf{Intercept}($\beta_0$) & \textbf{d($\beta_1$)} & \textbf{c($\beta_2$)} & \textbf{d:c($\beta_3$)} \\
       \hline
       \textbf{4-5} & 0.831 & 0.0005 & 0.879 & 0.0003$*$ \\
       \textbf{4-6} & 0.426 & 0.011 & 0.099 & 0.002 \\
       \textbf{4-7} & 0.295 & 0.592 & 0.017 & 0.086 \\
       \textbf{5-6} & 0.463 & 0.042 & 0.285 & 0.109 \\
       \textbf{5-7} & 0.387 & 0.001 & 0.113 & 0.006 \\
       \textbf{6-7} & 0.918 & 0.048 & 0.600 & 0.104 \\
       \hline
    \end{tabular}
\end{table}

\begin{figure}
    \centering
    \includegraphics[width=\linewidth]{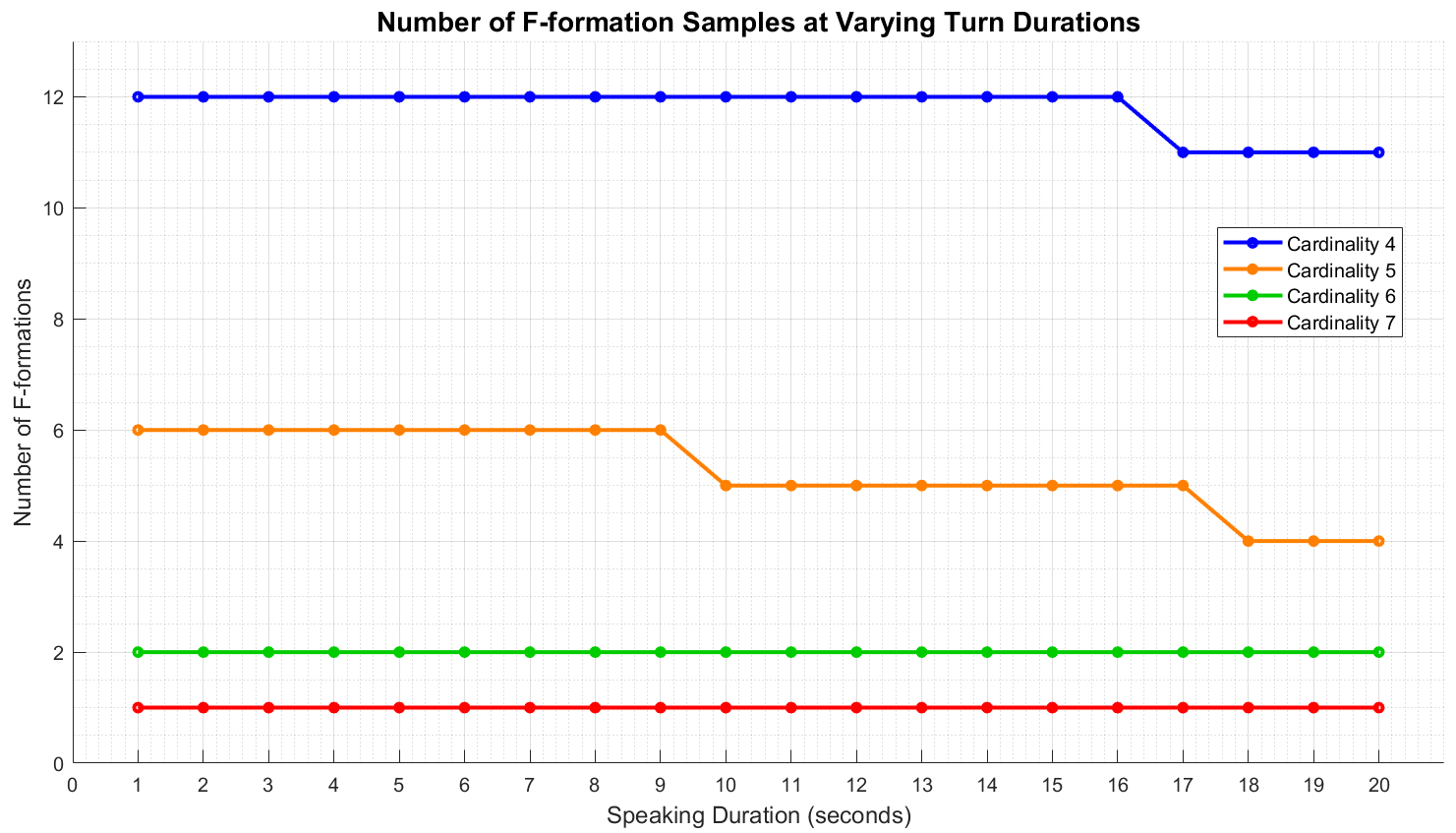}
    \caption{Number of F-formations at different speaking turn durations.}
    \label{fig:duration_num}
\end{figure}

We acknowledge that a possible limitation of this analysis is the difference and scarcity of the number of available F-formations under some circumstances. We present the number of F-formations per cardinality with respect to speaking turn duration in figure~\ref{fig:duration_num}.

\subsection{Spatial Analysis Results}


\begin{figure*}[!t]
    \centering
    \subfloat[]{
    \includegraphics[width=0.5\linewidth]{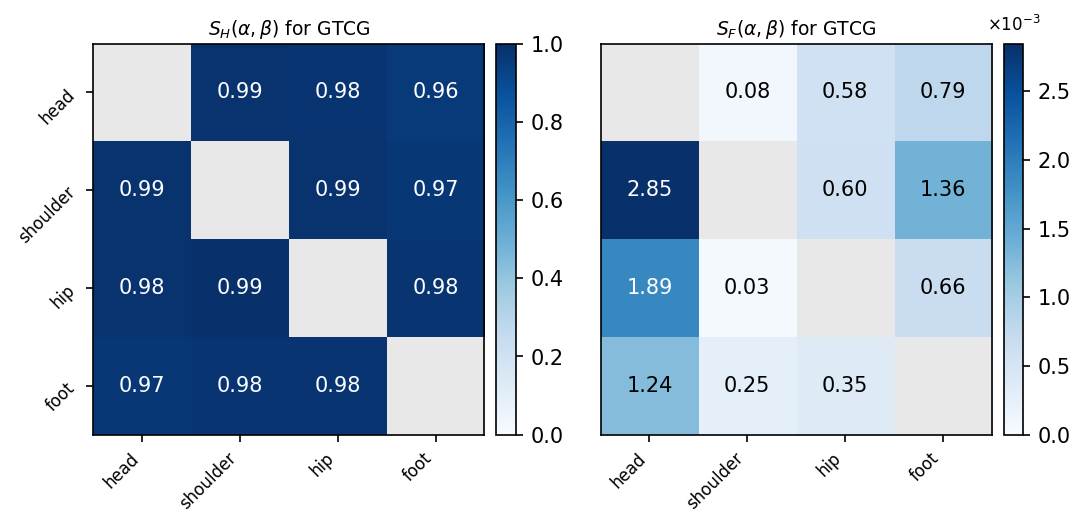} \label{fig:spatial_gtcg}}
    \subfloat[]{
    \includegraphics[width=0.5\linewidth]{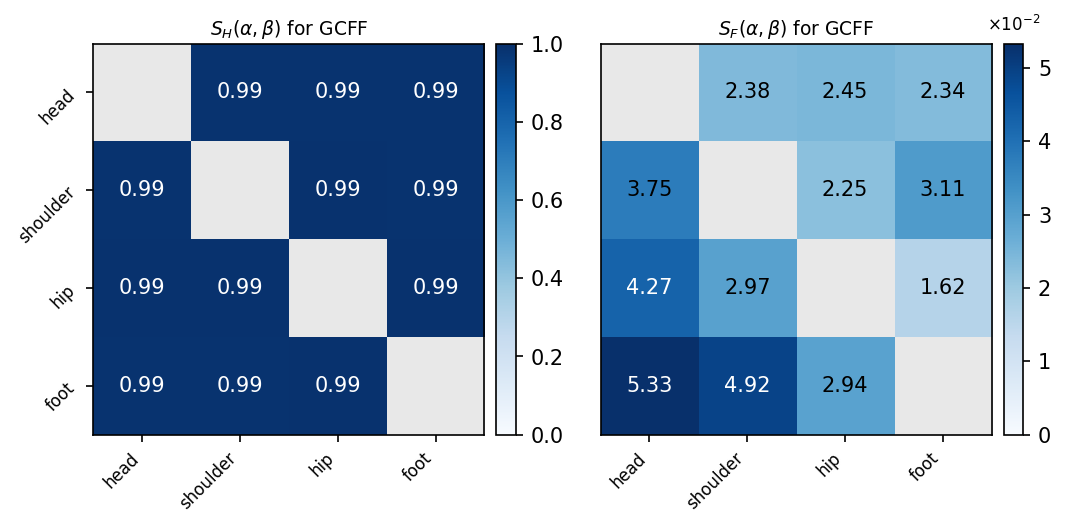}
    \label{fig:spatial_gcff}}
    \caption{Spatial analysis results for both pipelines. All figures are interpreted from row to column. For instance, a cell located at \textit{shoulder} row and \textit{hip} row of SH is interpreted as $S_H(\text{shoulder},\text{hip})$. ~\ref{fig:spatial_gtcg}: SH and SF for GTCG pipeline. SH is at least $0.96$ which shows high agreement over groupinf patterns. SF shows that it takes highest for the $(\text{shoulder},\text{head})$ pair and is substantially higher when $\beta$ takes head features. ~\ref{fig:spatial_gcff}: SH and SF for GCFF pipeline. SH score is $0.99$ for every pair of different $(\alpha, \beta)$, showing high agreement over groupings. SF shows clear asymmetry: $S_F(\alpha, \beta)$ is much larger than the opposite when $\alpha$ is physically lower than $\beta$ on the body.}
    \label{fig:spatial_analysis}
\end{figure*}


\subsubsection{\textbf{Detected groups have high agreement over different clues}} 
The SH and SF scores for both pipelines are shown in figure~\ref{fig:spatial_analysis} and our observation is similar for both pipelines: $S_H(\alpha, \beta)$ is close to 1 for every pair of clues. This indicates that running both pipelines would give us high agreement over the splitting pattern of subgroups: either one detection subsume the other, or they are nearly identical. It reveals that although the detection themselves are noisy across frames, they produce high agreement over groupings of people. Suppose with a frame of person set $P=\{1,2,3,4,5\}$ as in figure~\ref{fig:pipeline}, the differences in detection regarding to different body clues are more of a higher level of organization of smaller subgroups: all clues would mostly agree that $[1,2,3]$, and $[4,5]$ are within the same group, and only vary on whether to merge the two groups together to form a larger, coarser group, instead of a random restructuring on individual level. A complete shuffle, for example, one clue produces $[1,2,3], [4,5]$ while the other results in $[1,3,5], [2,4]$, is very unlikely. 

\subsubsection{\textbf{LBF detected results split into UBF groups, especially those from head features, more than inversely}} 
The SF scores tell us the direction of the splitting, as illustrated in section~\ref{sec:method}. In figure~\ref{fig:spatial_gtcg} and~\ref{fig:spatial_gcff}, we observe that SF show clear asymmetrical patterns. Specifically, for GTCG pipeline, $S_F(\alpha,\beta)$ takes the largest value when $(\alpha,\beta)=(\text{shoulder},\text{head})$. Also, SF is larger when $\beta$ takes head features. This shows that it happens more frequently when groups detected from other features split into groups detected from head features, indicating that more frequently people formulate smaller, finer-grained groups with positions and orientations of the head, while other parts remain in the larger group, especially with their shoulders. 
For GCFF pipeline, the observation is even stronger, as the SF is significantly larger in the bottom-left corner than the top-right corner. This means $S_F(\alpha, \beta)$ is larger when $\alpha$ belongs to LBF like hip or foot, and $\beta$ belongs to UBF like head or shoulder. $S_F(\alpha, \beta)$ takes the largest value for (foot, head) pair, and it is more than 2 times larger than its symmetrical pair of (head, foot). This means that statistically, groups detected from foot features split into groups resulting from head features more than 2 times than the inverse case. Also, we note that for each pair of $(\alpha, \beta)$, $S_F(\alpha, \beta) > S_F(\beta, \alpha)$ if $\alpha$ is physically lower than $\beta$, aligning with the the observed pattern.
We also notice that the SF is much smaller for GTCG than GCFF, as the bar scale for the former is $10^{-3}$ compared to the $10^{-2}$ for the latter. This indicates that the splitting over two proper groups happens much sparser in GTCG pipeline than GCFF.

Nevertheless, SF from UBF to LBF is still non-zero for both pipelines. This means at least in some cases, groups from UBF would also split into groups from LBF. This may be due to inaccurate keypoint extraction, internal feature of the pipeline itself, or the fact that sometimes people do orient their heads towards each other and formulate larger groups than their lower body configurations. However, our analysis reveals that this case is much rarer than the opposite, and in the following joint analysis we illustrate that in these cases it is less typical for the hypothesis that one F-formation could contain more than one floor.

\subsection{spatial-temporal joint analysis} \label{sec:cross_result_analysis}
\begin{figure*}[!t]
    \centering
    \subfloat[]{\includegraphics[width=0.5\linewidth]{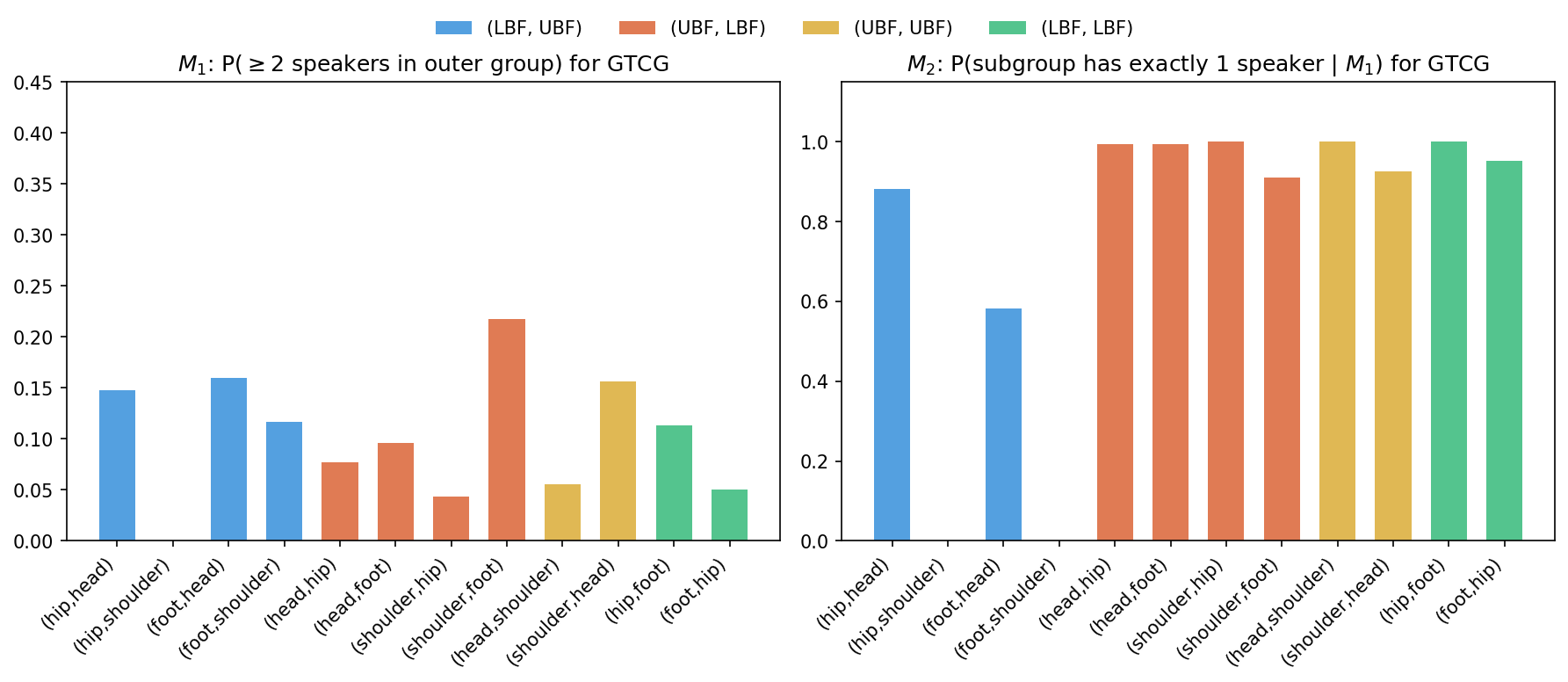}
    \label{fig:window_dist_gtcg}}
    \subfloat[]{\includegraphics[width=0.5\linewidth]{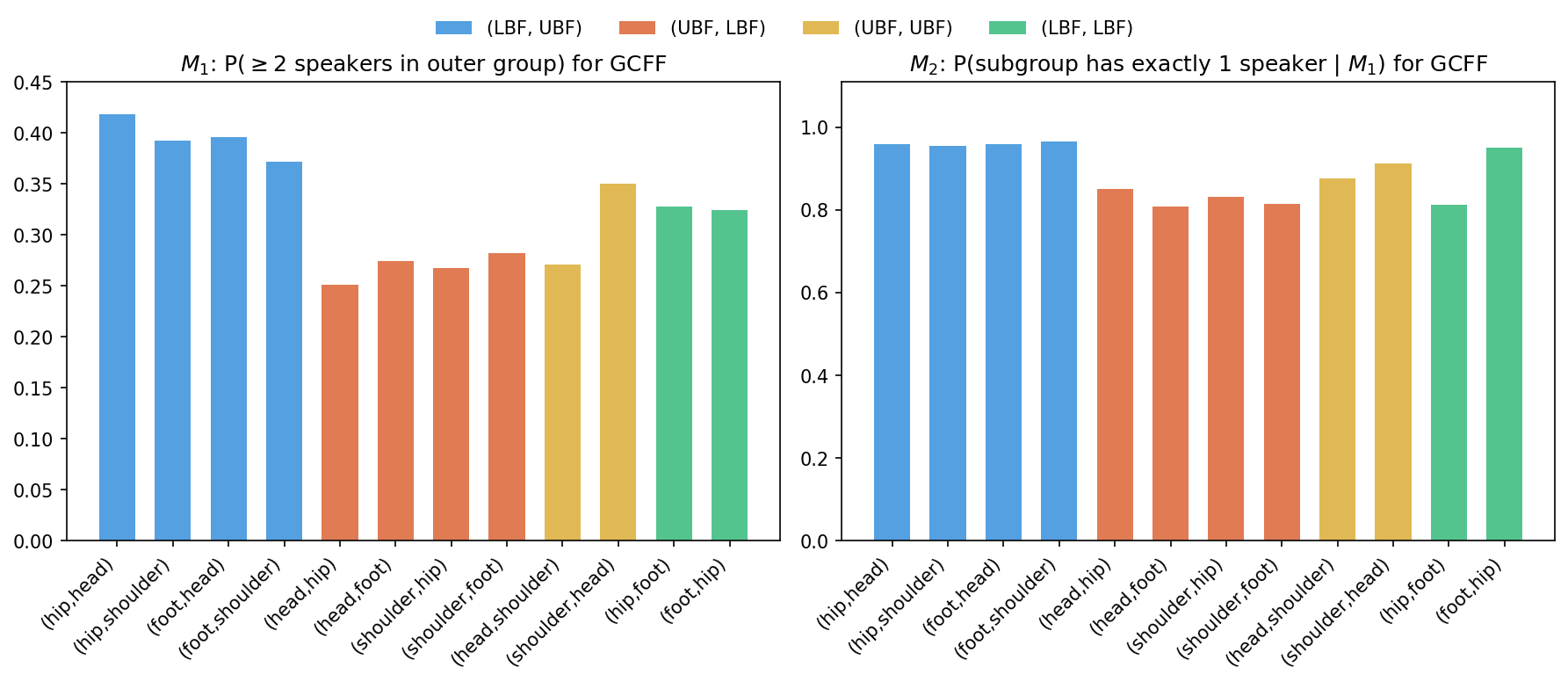}
    \label{fig:window_dist_gcff}}
    \caption{$M_1$ and $M_2$ for each clue pair for both pipelines. Within the labels, $\alpha \rightarrow \beta$ refers to $M(\alpha,\beta)$. $M_1$ calculates how often there are at least 2 speakers in the larger group when a split event happens, and $ M_2$ calculates how frequently each subgroup contain exactly one speaker when $M_1$ holds. The data bars are grouped in 4 sets: (LBF, UBF), (UBF, LBF), within UBF, and within LBF. ~\ref{fig:window_dist_gtcg}: $M_1$ and $M_2$ for GTCG. ~\ref{fig:window_dist_gcff}: $M_1$ and $M_2$ for GCFF.}
\end{figure*}

In this section we present results for the level of alignment between the subdivided groups and true conversation floors. We show the results for $\bar M_1$ and $\bar M_2$ in figure~\ref{fig:window_dist_gtcg} and~\ref{fig:window_dist_gcff} for the results of GTCG and GCFF respectively. 

\subsubsection{\textbf{Larger group containing more than one floor is generally more common when LBF groups split into UBF groups}}
The $M_1$ scores are showing similar trending for both pipelines. $M_1(\alpha, \beta)$ is larger when $\alpha$ belongs to LBF and $\beta$ belongs to UBF, except for the case of (shoulder, foot) and (hip,shoulder) in GTCG pipeline. This indicates that when groups from LBF do split into groups from UBF, the larger group contains more than one floor more frequently. Inversely, when UBF groups split into multiple LBF groups, this is less frequent. This indicates that the pattern captured in LBf groups splitting into UBF groups align more with the hypothesis that the larger F-formation contain more than one floor. Specifically for GTCG, the $M_1$ score is generally lower than GCFF. There are no instances for the (hip,shoulder) pair, and (shoulder, foot) pair has the highest value. Nevertheless, all other instances of (LBF, UBF) pair are higher than the opposite. For GCFF, the pattern is cleaner: $M_1(\alpha, \beta)$ is approximately $0.12$ higher when computed from LBF to UBF than the opposite direction. The $M_1$ score within LBF or UBF, is placed in between and in consistency with the observation.

\subsubsection{\textbf{Relationship between smaller groups and conversation floors is pipeline specific}}
The $M_2$ scores evaluates that when such a splitting happens, the proportion of subgroups which contain exactly one speaker. For GTCG, we observe that $M_2$ is higher for (UBF,LBF) pairs, and two pairs in (LBF, UBF) are zero, indicating that there is no such case where the subgroups contain exactly one speaker. Although this doesn't align with our assumptions that groups from UBF features correspond more to conversation floors, it could be influenced by the fact that these splits are sparse anyway in GTCG pipeline, as previously mentioned. GCFF pipeline, however, aligns with our hypotheses much more. We observe that when it is computed from LBf groups splitting into UBF groups, $M_2$ is close to $1$, indicating that whenever the larger group contain more than one speaker, almost all subgroups contain one speaker each. This is a strong indication that when the splitting pattern from LBF groups to UBF groups is observed, it is almost certain that each of the UBF groups contain exactly one floor. On the other hand, when UBF groups split into LBF groups and the larger group has more than one speaker, it is less certain that the LBF groups contain one speaker each, although the probability is still high. Combined together with the $M_1$ measurement, it demonstrates that when such splitting pattern occurs in GCFF, we observe much more instances for the case of LBF splitting into UBF groups that it is actually corresponding to the fact that one F-formation containing more than one floor. When detected UBF groups split into LBF groups, on the other hand, is less likely observed, and less likely resulting from such subsumption relationship.

\section{Conclusion} \label{sec:conclusion}

In the above experiments, we have studied the difference of the F-formation detection results from different feature sets. Summing up the observations, we conclude that
\begin{itemize}
    \item An F-formation may contain more than one conversation floor in both theoretical and realistic aspects;
    \item For existing pipelines for detecting F-formations, using different clues from body parts makes a difference. Some resulting groups using LBF split into two or more in groups resulted from UBF and this happens more frequently than the opposite splitting, especially when groups from other features splits into groups from head features. This indicates that head positions and orientations may carry more fine-grained conversational involvement information within a larger interaction context.
    \item When LBF groups split into UBF groups, the correspondence between larger/smaller groups is pipeline specific. In GCFF pipeline, we observe clear alignment that LBF groups correspond to F-formations and UBF groups correspond to single conversation floors. This indicates that certain methods could be promising in detecting and translating conversational involvement at floor level.
\end{itemize}

\section{Discussion and Limitation}
\noindent \textbf{Discussion.} The results in this article show that current operationalization of FCGs is not enough for capturing single floors. We propose the following points for future interests:

\begin{itemize}
    \item Develop a multi-tier theory to describe involvement in FCGs. Operationalizing FCGs as F-formations are insufficient to encapsulate single conversation floors. However, it does not mean that F-formations does not carry any socially meaningful information regarding involvement. By defining it with the lower-body orientations, it still suggests some structure in which common activities take place - or common interest/attention in cognitive perspective, but this structure may contain multiple conversation floors. Therefore, we propose that a multi-tier theory may be of interest to future studies, which could describe different categories of interactional involvement on different levels, and F-formations/conversation groups correspond to one particular level/tier.
    \item A more fine-grained model to capture VFoA of people. With the advance in skeleton annotation in multi-person scenes, we believe that VFoA is not just about the direction of eye gaze or body pose, but a more complicated structure which integrates features from all around the body. Using a single orientation may overlook certain aspects in attention, and describing such attention mechanism in multi-dimensional way may pose further discoveries regarding involvement in an interaction.
\end{itemize}

\noindent \textbf{Limitation.} We used an approximation of using binary speaking status to account for conversation floors. Although this is conceptually feasible as in our arguments, it may not afford identically towards the analysis of floors as real speech data. We made this design choice to be privacy preserving, but future work may reveal different patterns for floors if privacy is not an issue. Also, we leverage keypoint data to infer positions and orientations of people (see supplemental material for details), which may not be as accurate as datasets with full mesh capturing and allows for direct feature extraction. 

Our conclusion should not be interpreted as exact correspondence between body feature and F-formation/floor distinction. Rather, we point out the statistical difference in the group patterns that result from different features. The exact profile of correspondence should be figured out by future works.

\section*{Acknowledgments}
We thank Gara Dorta Perez for explorations of feature extraction process in the ConfLab dataset. Plotting of figure~\ref{fig:pipeline} and~\ref{fig:joint_analysis_pipeline} as well as this paper's code implementation~\footnote{http://claude.ai/design; http://claude.ai/code} is done with AI assistance. 



\printbibliography


\section{Biography Section}

\vspace{-11pt}

\begin{IEEEbiographynophoto}{Zonghuan Li} received BEng degree in Electrical Engineering from Tsinghua University in China, 2016, and master's degree in Electrical Engineering from University of California, San Diego in United States, 2023. He is currently a PhD student under the Socially Perceptive Computing Lab, TU Delft, Netherlands since 2024. His research interests include conversational involvement analysis with multimodal signals, multimodal large language model applications and evaluation. 
\end{IEEEbiographynophoto}

\vspace{-11pt}
\begin{IEEEbiographynophoto}{Chirag Raman} received his master’s degree in entertainment technology from Carnegie Mellon University, USA in 2013 and the Ph.D in Computer Science from TU Delft in 2023. He is currently an Assistant Professor at TU Delft where he leads the Tapri Lab within the Department of Intelligent Systems. His research interests include gnerative modeling, socially intelligent embodied agents, with insights from cognitive science and social psychology. 
\end{IEEEbiographynophoto}

\vspace{-11pt}
\begin{IEEEbiographynophoto}{Hayley Hung} (Member, IEEE) received the PhD degree in computer vision from the Queen Mary University of London in 2007. She is currently an Associate Professor with the Socially Perceptive Computing Lab, TU Delft, The Netherlands, where she works since 2013. She works towards finding how automated systems best serve people with social awareness. Her research interests include devising multimodel models to automatically interpret face-to-face human social behavior. 
\end{IEEEbiographynophoto}

\vfill
\clearpage

\section*{Appendix: Feature extraction of ConfLab dataset}





Here we introduce the process to obtain positions and orientations from different body parts from the ConfLab~\cite{raman2022conflab} dataset. Specifically, for each person in a frame and for each body feature $\alpha \in \mathcal{C}=\{\text{head,shoulder,hip,foot}\}$, we extract positions $\mathbf{p}_\alpha$ and orientations $\theta_\alpha$, which are then processed by GTCG and GCFF pipelines. This comprises three steps. 

\subsection{Extracting 2D keypoints}
To infer positions and orientations, we use keypoints as an anchor. ConfLab provides video recording from top-view cameras at 60 fps. We first use SAM3~\cite{carion2025sam} model to obtain a mask for each person, and then use ViTPose~\cite{xu2022vitpose} to obtain 2D keypoints for them, by providing the convex hull of each mask as the bounding box. This results in 2D keypoints in COCO format~\footnote{https://cocodataset.org/\#home} for each person and each frame. It contains head, pose, and shoulder/hip/ankle/toe keypoints on left and right sides. 

The 2D keypoints are denoted by $\mathbf{x} \in \mathbb{R}^2$ which lies in the pixel domain. In the following context, subscripts \textit{h}, \textit{n} \textit{s}, \textit{i}, \textit{a}, \textit{t} stands for head, nose, shoulder, hip, ankle, and toe respectively. We also use $l$, $r$ subscripts to denote left/right keypoint, which applies to all keypoints other than head and nose. For instance, $\mathbf{x}_{l,s}$ represents pixel coordinates of the left shoulder keypoint.


\begin{figure*}[!t]
    \centering
    \includegraphics[width=\linewidth]{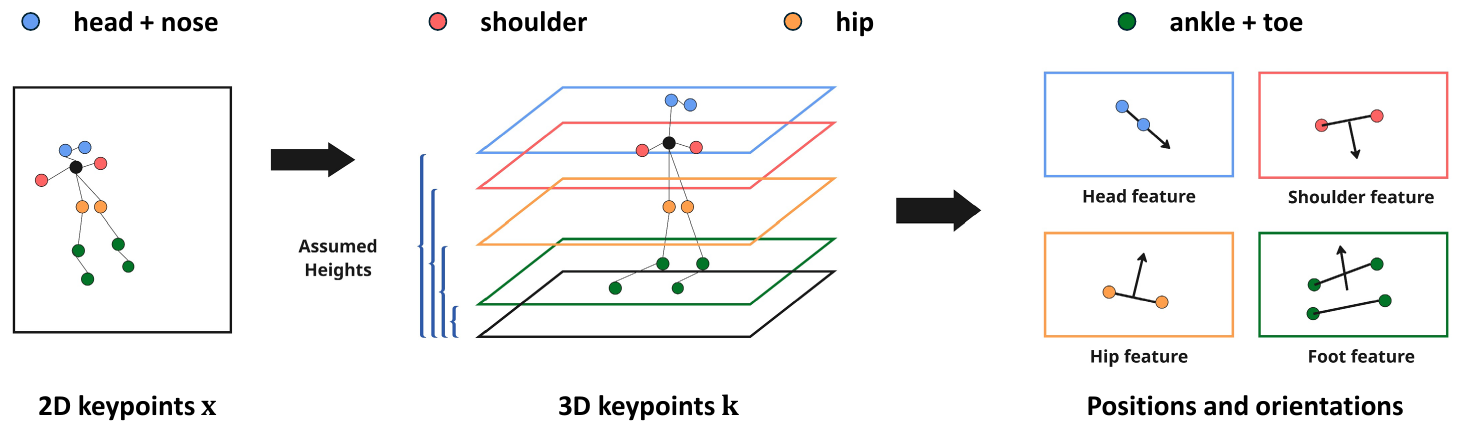}
    \caption{Illustration of the position and orientation extraction pipeline. Left: we construct 2D keypoints on a frame in COCO format; Middle: we project the keypoint into 3D space by assuming the height of each keypoint; Right: we construct for sets of features (positions and orientations) to leverage in GTCG/GCFF pipelines.}
    \label{fig:kp_pipeline}
\end{figure*}
\subsection{3D Projection of keypoints}
2D keypoints on pixel domain is insufficient for GTCG and GCFF pipelines, as these methods assume positions and orientations to be on a plane in 3D space, instead of in 2D pixel space. Therefore, we transfer the keypoints to 3D by leveraging intrinsic matrix $K$ and extrinsic matrix $C$ provided by the ConfLab dataset. We do this by assuming the heights of each keypoint, and calculate its space coordinates by taking the the intersection of the ray represented by the pixel keypoint and the horizontal plane at the height level.

Specifically, since we need head, shoulder, hip, ankle, and toe keypoints, we assume that the height for each keypoint is $h_{j} = \gamma_j h$, where $j$ represents the feature and $j\in \mathcal{K} = \{ \text{head,nose,shoulder,hip,ankle,toe} \}$. $h$ is the assumed common height of a person. $\gamma$ is a height coefficient that indicates the ratio between the respective keypoint and the total body height. In practice we use $h=1.7$m and we show the assumed $\gamma$ in table~\ref{tab:heights}. This process is shown in the left part of figure~\ref{fig:kp_pipeline}.

After this step we obtain keypoint coordinates in 3D space $\mathbf{k}_j$ where $j\in \mathcal{K}$.

\subsection{Body-part specific feature construction}

After transferring the keypoints into 3D, we illustrate how the the input features are constructed. We need to note here that although we have transferred into the 3D space and the coordinates are in $\mathbb{R}^3$, we omit the third dimension and limit the discussion to 2D in this context, i.e. on the plane that is parallel to the ground, and the third coordinate remain the identical height as in previous step. Since both GTCG and GCFF pipelines take only 2D inputs, this is unharmful. Therefore, all $\mathbf{p}$s and $\mathbf{k}$s in the following context are in $\mathbb{R}^2$. 

From the previous step we obtain keypoints in space as $\mathbf{k}_j \in \mathbb{R}^2$. We now construct a set of positions $\mathbf{p}$ and orientations $\theta$ as follows:
\begin{equation}
\begin{split}
    & \mathbf{p}_h = \mathbf{k}_h; \\
    & \mathbf{p}_s = (\mathbf{k}_{l,s} + \mathbf{k}_{r,s}) / 2; \\
    & \mathbf{p}_i = (\mathbf{k}_{l,i} + \mathbf{k}_{r,i}) / 2; \\
    & \mathbf{p}_f = (\mathbf{k}_{l,a} + \mathbf{k}_{r,a} + \mathbf{k}_{l,t} + \mathbf{k}_{r,t}) / 4; \\
\end{split}
\end{equation}

\begin{equation}
\begin{split}
    & \theta_h' = \arg(\mathbf{k}_n - \mathbf{k}_h); \\
    & \theta_s' = \arg(\mathbf{R}_{\frac{\pi}{2}}(\mathbf{k}_{r,s} - \mathbf{k}_{l,s})); \\
    & \theta_i' = \arg(\mathbf{R}_{\frac{\pi}{2}}(\mathbf{k}_{r,h} - \mathbf{k}_{l,h})); \\
    & \theta_f' = (\theta_a' + \theta_t') / 2, \quad \text{where} \\
    & \quad \theta_a' = \arg(R_{\frac{\pi}{2}}(\mathbf{k}_{r,a} - \mathbf{k}_{l,a})) \\
    & \quad \theta_t' = \arg(R_{\frac{\pi}{2}}(\mathbf{k}_{r,t} - \mathbf{k}_{l,t}))
\end{split}
\end{equation}

For subscripts, $h,s,i,f$ denotes head, shoulder, hip, and foot respectively. Note that the foot feature is an integration of ankle and toe keypoints.
For the orientations, assume $\mathbf{v} = (x,y) \in \mathbb{R}^2$, then $\arg(\mathbf{v}) = \tan^{-1}(y/x)$ calculates the argument of a vector in terms of radians and ensures it to be within $[0,2\pi)$, and 
\begin{equation}
    R_{\phi} = \begin{bmatrix}
        \cos \phi & -\sin \phi \\ 
        \sin \phi & \cos \phi 
    \end{bmatrix}. 
\end{equation}
is the rotation matrix. For shoulder and hip orientations, the above operation is equivalent to calculating the orientation of the normal vector of the segment which points from the left keypoint to the right one. For foot orientations, it takes the mean value of the orientation obtained by doing the same operation to ankle and toe.

Finally, each set of constructed $(\mathbf{p}, \theta)$ for all person and all frames are then used in the GTCG/GCFF pipelines.

\begin{table}[!t]
    \caption{Choice of height coefficient}
    \label{tab:heights}
    \centering
    \begin{tabular}{|c||c|c|c|c|c|c|}
    \hline
       Keypoint & head & nose & shoulder & hip & ankle & toe \\
       \hline
       $\gamma$ & 1 & 0.95 & 0.85 & 0.5 & 0.02 & 0.02 \\
       \hline
    \end{tabular}
\end{table}



\end{document}